\documentstyle[11pt,pstricks,pst-node,eaclap]{article}
\input psfig

\title{\vspace{-0.5in}A Robust Parser Based on  Syntactic Information}
\author{Kong Joo Lee \hspace{0.5cm}  Cheol Jung Kweon \hspace{0.5cm} Jungyun Seo \hspace{0.5cm} Gil Chang Kim\\
Department of Computer Scinence and CAIR\\
Korea Advanced Institute of Science and Technology\\
Taejon, Korea  305-701\\
\{kjlee,cjkwn\}@csone.kaist.ac.kr\\}

\begin{document}

\maketitle

\vspace{-0.5in}
\begin{abstract}

An extragrammatical sentence is what a normal parser fails to analyze. 
It is important to recover it using only syntactic information
although results of recovery are better if semantic factors are considered.
{\it A general algorithm for least-errors recognition},  
which is based only on syntactic information, 
was proposed by G. Lyon to deal with the extragrammaticality. 
We extended this algorithm to recover extragrammatical sentence into grammatical one
in running text.
Our robust parser with recovery mechanism
-- extended general algorithm for least-errors recognition --
can be easily  scaled up and modified because it utilize only
syntactic information.
To upgrade this robust parser
we proposed heuristics through the analysis on the Penn treebank corpus.
The experimental result shows 68\% $\sim$ 77\% accuracy in error recovery.  
\end{abstract}

\section{Introduction}

Extragrammatical sentences include patently ungrammatical constructions as well as
utterances that may be grammatically acceptable but are beyond the syntactic coverage
of a parser, and any other difficult ones that are encountered in parsing \cite{def}.

\vspace{7pt}
{\small\sf
\setlength{\baselineskip}{10pt}
	\begin{tabbing}
	aa\=\kill
	\>I am sure this is what he means.\\
	\>This is, I am sure, what he means.\\
	\> \\
	\>The progress of machine does not stop even a day.\\
	\>Not even a day does the progress of machine stop.\\
	\end{tabbing}
	\vspace{-5mm}
	 
}
\vspace{7pt}

Above examples show that people are used to write same meaningful sentences differently.
In addition, people are prone to mistakes in writing sentences.
So, the bulk of written sentences are open to the extragrammaticality.
\par
In the Penn treebank~ tree-tagged ~corpus\cite{marcus}, for instance, about 80 percents of the rules are
concerned with peculiar sentences which include
inversive, elliptic, parenthetic, or emphatic phrases.
For example, 
we can drive a rule {\it VP  $\rightarrow$  vb  NP  comma  rb  comma  PP} ~~from the following sentence.

\vspace{7pt}
{\sf
\begin{quote}
The same jealousy can breed confusion, {\it however,} in the absence of any authorization bill this year.
\end{quote}
}

	{\small
	\setlength{\baselineskip}{10pt}
	\begin{verbatim}
(
(S
 (NP The/dt
  (ADJP same/jj) jealousy/nn) can/md
 (VP breed/vb
  (NP confusion/nn) ,/, however/rb ,/,
  (PP in/in
   (NP
    (NP the/dt absence/nn)
    (PP of/in
     (NP any/dt authorization/nn bill/nn))
    (NP this/dt year/nn)))))
./.)

	\end{verbatim}
	}

A robust parser is one that can analyze these extragrammatical sentences without failure. 
However, if we try to preserve robustness 
by adding such rules whenever we encounter an extragrammatical sentence,
the rulebase will grow up rapidly, and thus processing and maintaining the excessive 
number of rules will become inefficient and impractical.
Therefore, extragrammatical sentences should be handled by some recovery mechanism(s)
rather than by a set of additional rules.
\par
Many researchers have attempted several techniques to deal with extragrammatical sentences
such as Augmented Transition Network(ATN) \cite{atn},
network-based semantic grammar \cite{sg}, partial pattern matching \cite{ppm},
conceptual case frame \cite{cf}, and  
multiple cooperating methods \cite{mm}.
Above mentioned techniques take into account 
various semantic factors 
depending on specific domains on question
in recovering extragrammatical sentences. 
Whereas they can provide even better solutions intrinsically,
they are usually ad-hoc and  are lack of extensibility.
Therefore, it is important to recover extragrammatical sentences
using syntactic factors only, which are independent of any particular system
and any particular domain.
\par
Mellish \cite{mell} introduced some chart-based techniques using only syntactic
information for extragrammatical sentences.
This technique has an advantage that there is no repeating work 
for the chart to prevent
the parser from generating the same edge as the previously existed edge.
Also, because the recovery process runs when a normal parser terminates 
unsuccessfully, the performance of the normal parser does not decrease
in case of handling grammatical sentences.
However, his experiment was not based on the errors in running texts
but on artificial ones which were randomly generated by human.
Moreover, only one word error was considered though
several word errors can occur simultaneously in the running text.

\par
{\it A general algorithm for least-errors recognition} \cite{gl}, proposed by G. Lyon, 
is to find out the least number of errors necessary to successful parsing and
recover them.
Because this algorithm is also syntactically oriented and based on a chart, it has the same
advantage as that of Mellish's parser.
When the original parsing algorithm terminates unsuccessfully, the algorithm begins to
assume errors of insertion, deletion and mutation of a word.
For any input, including grammatical and extragrammatical sentences, 
this algorithm can generate the resultant parse tree.
At the cost of the complete robustness, however, this algorithm degrades the
efficiency of parsing, and generates many intermediate edges.

\par
In this paper, we present a robust parser with a recovery mechanism.
We extend {\it the general algorithm for least-errors recognition}
to adopt it as the recovery mechanism in our robust parser.
Because our robust parser handle extragrammatical sentences with  
this syntactic information oriented recovery mechanism,
it
can be independent of a particular system or particular domain.
Also, we present the heuristics to reduce the number of edges
so that we can upgrade the performance of our parser.

\par
This paper is organized as follows :
We first review a general algorithm for least-errors recognition.
Then we present the extension of this algorithm, and 
the heuristics adopted by the robust parser. 
Next, we describe the implementation of the system and the result of the experiment of
parsing real sentences. 
Finally, we make conclusion with
future direction.

\section{Algorithm and Heuristics}

\subsection{General algorithm for least-errors recognition}
\label{sec:2}

\par
The general algorithm for least-errors recognition \cite{gl}, 
which is based on Earley's algorithm,
assumes that sentences may have insertion, deletion, and
mutation errors of terminal symbols.
The objective of this algorithm is to parse
input string with the least number of errors.

\par
A {\it state} used in this algorithm is quadruple {\it (p, j, f, e)}, where
{\it p} is a production number in grammar, 
{\it j} marks a position in {\it RHS(p)},
{\it f} is a start position  of the state in input string, and
{\it e} is an error value.\footnote{Lyon said that e is an error count}
A {\it final state (p, $\underline{p}$+1, f, e)} denotes recognition of a phrase
{\it RHS(p)} with {\it e} errors where $\underline{p}$ is a number of components in rule {\it p}.
A {\it stateset S(i)}, where {\it i} is the position of the input, is an ordered set of states.
States within a stateset are ordered by ascending value of {\it j},
within a {\it p} within a {\it f}~; {\it f} takes descending value.

When adding to statesets, if state {\it (p, j, f, e)} is a candidate for admission 
to a stateset which already has a similar member {\it (p, j, f, e')} and
{\it e'} $\leq$ {\it e}, then {\it (p, j, f, e)} is rejected.
However, if {\it e'} $>$ {\it e}, then {\it (p, j, f, e')} is replaced
by {\it (p, j, f, e)}.

The algorithm works as follows :
A procedure SCAN is carried out for each state in {\it S(i)}.
SCAN checks various correspondences of input token {\it t(i)} against 
terminal symbols in {\it RHS} of rules.
Once SCAN is done, COMPLETER substitutes all final states of {\it S(i)}
into all other analyses which can use them as components.

\begin{figure}
   
	\centerline{\psfig{figure=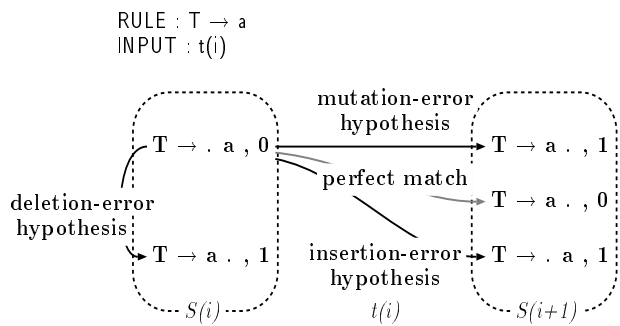,height=4.4cm,width=8.3cm}}
	\caption{SCAN processing}
	\label{fig1}
    \end{figure}

\noindent {\bf SCAN}\\
SCAN handles states of {\it S(i)}, checking each input terminal against requirements 
of states in {\it S(i)} and various error hypotheses.
Figure~\ref{fig1} shows how SCAN processes.

Let {\it c(p,j)} be {\it j-th} component of {\it RHS(p)} and
{\it t(i)} be {\it i-th} word of input string.
\begin{itemize}
\item	{\it perfect match} :	\\
		If {\it c(p,j)} = {\it t(i)} then add {\it (p, j+1, f, e)} to 
		{\it S(i+1)} if possible. 
\item	{\it insertion-error hypothesis} : \\
		Add {\it (p, j, f, e+$\alpha_{insertion}$)} to {\it S(i+1)} if possible.\\
        $\alpha_{insertion}$ is the cost of an insertion-error for a terminal symbol.
\item	{\it deletion-error hypothesis} :\\
		If {\it c(p,j)} is terminal, then add {\it (p, j+1, f, e+$\alpha_{deletion}$)} 
		to {\it S(i)} if possible. \\
        $\alpha_{deletion}$ is the cost of a deletion-error for a terminal symbol.
\item	{\it mutation-error hypothesis} :	\\
		If {\it c(p,j)} is terminal but not equal to {\it t(i)},
		then add {\it (p, j+1, f, e+$\alpha_{mutation}$)} to {\it S(i+1)} if possible.\\
        $\alpha_{mutation}$ is the cost of a mutation-error for a terminal 
        symbol.\footnote{$\alpha_{insertion},~\alpha_{deletion},~\alpha_{mutation}$ are all strictly
        1 in Lyon's original paper.}
\end{itemize}

\noindent {\bf COMPLETER}\\
COMPLETER handles substitution of final states in {\it S(i)} 
like that of original Earley's algorithm.
Each final state means the recognition of a nonterminal.

\subsection{Extension of least-errors recognition algorithm}

The algorithm in section~\ref{sec:2} can analyze
any input string with the least number of errors.
But this algorithm 
can handle only the errors of terminal symbols
because it doesn't consider the errors of nonterminal nodes.
In the real text, however, the insertion, deletion, or inversion of a
phrase  --  namely, nonterminal node --  
occurs more frequently.
So, we extend the original algorithm in order to handle the errors of 
nonterminal symbols as well.

\par
\begin{figure}
	\centerline{\psfig{figure=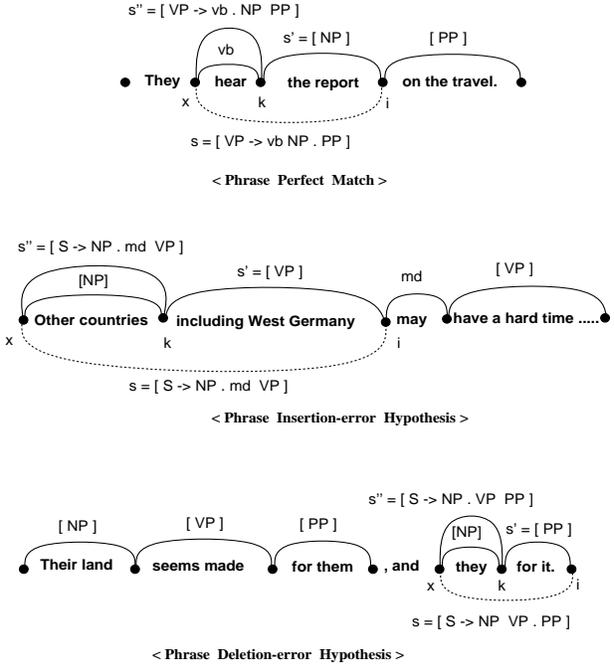,height=8.9cm,width=8.5cm}}
	\caption{Examples of extended-COMPLETER processing}
	\label{complete}
	\end{figure}

In our extended algorithm, the same SCAN as that of the original algorithm
is used,
while COMPLETER is modified and extended.
Figure~\ref{complete} shows the processing of extended-COMPLETER.
In figure~\ref{complete}, {\it [NP]} denotes the final state 
whose rule has {\it NP} as its {\it LHS}.
In other words, it means the recognition of a noun phrase.

\noindent {\bf extended-COMPLETER}\\
	If there is a final state ${\it s' = (p', \underline{p'}+1, k, e')}$ in $S(i)$,	
\begin{itemize}
\item	{\it phrase perfect match}\\
		If there exists a state ${\it s'' = (p, j, x, e)}$ in $S(k)$ , $k < i$  and
		$c(p,j) = LHS(p')$
		then add $s = (p, j+1, x, e+e')$  into $S(i)$.

\item	{\it phrase insertion-error hypothesis
\footnote{In fact, there are cases 
that an inserted phrase cannot be constructed to form a nonterminal node. 
In phrase insertion-error hypothesis of figure~\ref{complete}, 
the original  sentence is ``Other countries, including West Germany, may have $\ldots$'', 
where the inserted phrase VP is surrounded by commas.
So, the substring{\it ( comma VP comma )} should be dealt with as a constituent
in extended-COMPLETER.
In fact, we implemented the algorithm to allow substring insertions as well as
insertions of nonterminal nodes.}
		}\\
		If there exists a state $s'' = (p, j, x, e)$ in $S(k)$ 
		then add $s = (p, j, x, e+\beta_{insertion})$ into $S(i)$ if possible.\\
		$\beta_{insertion}$ is the cost of a insertion-error for a nonterminal symbol.

\item	{\it phrase deletion-error hypothesis}\\
		If there exists a state $s'' = (p, j, x, e)$ in $S(k)$ 
		and $c(p,j)$ is a nonterminal
		then add $s = (p, j+1, x, e+\beta_{deletion})$ into $S(k)$ if possible.\\
		$\beta_{deletion}$ is the cost of a deletion-error for a nonterminal symbol.

\item 	{\it phrase mutation-error hypothesis}
		\footnote{We know that the phrase mutation-error hypothesis is not meaningful in the real text
			because we cannot find out any example of phrase mutation-error in the corpus.
		  So we didn't implement the phrase mutation-error hypothesis.}\\
		If there exists a state $s'' = (p, j, x, e)$ in $S(k)$ and 
		$c(p,j)$ is a nonterminal but  not equal to $L(p')$
		then add $s = (p, j+1, x, e+\beta_{mutation})$ into $S(i)$ if possible.\\
		$\beta_{mutation}$ is the cost of a mutation-error for a nonterminal symbol.
\end{itemize}

The extended least-errors recognition algorithm can handle 
not only terminal errors but also nonterminal errors.

\subsection{Heuristics}

\par
The robust parser using the extended least-errors recognition algorithm
overgenerates many error-hypothesis edges during parsing process.
To cope with this problem, we adjust error values according to the following
heuristics.
Edges with more error values are regarded as less important ones,
so that those edges are processed later than those of less error values.

\begin{itemize}
\item	{\bf Heuristics 1: error types}\\
		The analysis on 3,538 sentences of the Penn treebank corpus WSJ
		shows that there are 498 sentences with phrase deletions and
		224 sentences with phrase insertions.
		So, we assign less error value to the deletion-error hypothesis edge than 
		to the insertion- and  mutation-errors.

		\begin{tabbing}
		aa\=aaa\=aaa\=aaa\=\kill
		\>$\alpha$\>$<$\>$\beta$
		\end{tabbing}
		\begin{tabbing}
		aa\=aaaaaaaa\=aaa\=aaaaaaaaa\=aaa\=aaaaaaaaa\=\kill
		\>$\alpha_{deletion}$\>  $<$ \>    $\alpha_{insertion}$ \> $<$ \> $\alpha_{mutation}$\\
		\>$\beta_{deletion}$ \> $<$ \> $\beta_{insertion}$  \\
		\end{tabbing}

		where  $\alpha$ is the error cost of a  terminal symbol, $\beta$ is the error
	    cost of a nonterminal symbol. 

\item   {\bf Heuristics 2: fiducial nonterminal}\\
        People often make mistakes in writing English.
        These mistakes usually take place rather between small constituents
        such as a verbal phrase, an adverbial phrase and noun phrase
        than within small constituents themselves.
        The possibility of error occurrence within noun phrases are lower
        than between a noun phrase and a verbal phrase, a preposition phrase, an adverbial
        phrase. So, we assume some phrases, for example noun phrases,  as  fiducial nonterminals, which 
        means error-free nonterminals. When handling  sentences, the robust
        parser assings more error values($\delta_{1}$) to the error hypothesis edge 
        occurring within a fiducial nonterminal.

\item	{\bf Heuristics 3: kinds of terminal symbols}\\
        Some terminal symbols like punctuation symbols,
        conjunctions and particles are often misused.
        So, the robust parser assigns less error values($-\delta_{2}$) to the error hypothesis 
        edges with these symbols than to the other terminal symbols.

\item	{\bf Heuristics 4: inserted phrases between commas or parentheses}\\
		Most of inserted phrases are surrounded by commas or parentheses. For example,

		\vspace{8pt}
		{\small\sf
		\begin{description}
		\item[ a.]	They're active {\it , generally  ,} at night or on damp, cloudy days.
		\item[ b.]   All refrigerators {\it , whether they are defrosted manually or not ,} need to be cleaned.
		\item[ c.]  	I was a last-minute {\it ( read interloping )} attendee at a French journalism convention $\cdots$
		\end{description}
		}
		\vspace{8pt}

		We will assign less error values($-\delta_{3}$) to the insertion-error hypothesis edges 
		of nonterminals  which are embraced by  comma or parenthesis.\\
\end{itemize}
		$\delta_{1}$ and $\delta_{2}$ are weights for the  error of terminal nodes,
		and $\delta_{3}$ is a weight for the error of nonterminal nodes.

\par
The error value {\it e} of an edge is calculated as follows.
All error values are additive.\\
	The error value {\it e} for a rule  $ X \rightarrow a_{1} A_{1} a_{2} \cdots a_{i} A_{j} $,
	where {\it a} is a terminal node and {\it A} is a nonterminal node, is

	\begin{enumerate}
	\item 	$e  =  \sum_{1}^{i} e_{T} + \sum_{1}^{j} e_{NT} $

	\item	$e_{T} = \left\{ \begin{array}{ll}
				   			   \alpha + \delta_{1} - \delta_{2}  & \mbox{if terminal error}\\
				   				0								 & \mbox{otherwise}
				   \end{array}
				   \right. $ 
	\item	 $ e_{NT} = \left\{ \begin{array}{ll}
						     	 \beta -  \delta_{3}  + e_{child}  & \mbox{if nonterminal}\\
						     	 								   & \mbox{error}\\
				      e_{child}	  		 						    & \mbox{otherwise}
				      \end{array}
				      \right. $ 
	\end{enumerate}
where $\alpha \in \{\alpha_{insertion},~\alpha_{deletion},~\alpha_{mutation}\}$,
$\beta \in \{\beta_{insertion}, \beta_{deletion}\}$ and
$e_{child}$ is an error value of a child edge.

\par
By these heuristics, our robust parser can process only plausible edges first, 
instead of processing all generated edges at the same time,
so that  we can enhance the performance of the robust parser and result in
the great reduction in the number of resultant trees.

\section{Implementation and Evaluation}

\subsection{The robust parser}

\par
\begin{figure}[h]
\centerline{\psfig{figure=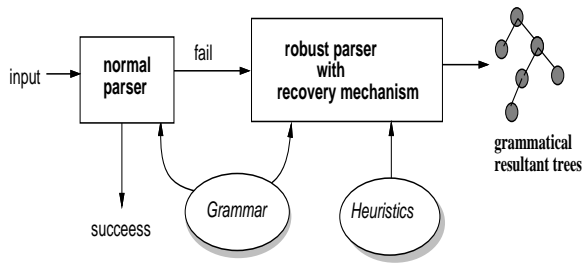,height=3.55cm,width=8.15cm}}
\caption{The overview of the system}
\label{fig2}
\end{figure}

Our robust parsing system is composed of two modules.
One module is a normal parser
which is the bottom-up chart parser.
The other is a robust parser with the error recovery mechanism proposed herein. 
At first, an input sentence is processed by  the normal parser.
If the sentence is within the grammatical coverage of the system,
the normal parser succeed to analyze it.
Otherwise, the normal parser fails, and then the robust parser starts to execute
with edges generated by the normal parser.
The result of the robust parser is the parse trees which are within the grammatical 
coverage of the system.
The overview of the system is shown in  figure~\ref{fig2}.

\subsection{Experimental result}

\par
To show usefulness of the robust parser proposed in this paper, 
we made some experiments.

\begin{itemize}

        \begin{table*}
        \begin{center}
        \caption{The results of the robust parser on WSJ}
        \label{tab1}
        \renewcommand{\arraystretch}{1.1}
        \begin{tabular}{|l||l|l|}\hline
        \multicolumn{3}{|c|}{Experiment 1 : WSJ  410 sentences} \\\hline
	                                &   with Heuristics             & without Heuristics\\\hline\hline
        Average sentence length     &   16.27 words (2-25 words)    & 16.27 words (2-25 words)   \\
        Average processing time     &   6.52 sec                  & 22.47 sec                \\
        Average number of edges     &   7726.03                   & 10346.6                   \\
        Accuracy (\%)         &   77.1                    & 72.8                  \\
        no-crossing  sentences   &   23.28\%         & 20.28\%        \\
        \% of $\leq$ 1-crossing sentences    &  40.52\% & 37.14\%       \\
        \% of $\leq$ 2-crossing sentences    &  55.17\% & 48.57\%      \\  \hline
        \end{tabular}
        \end{center}
        \end{table*}

	\item	{\it Rule}\\
			We can derive 4,958 rules and their frequencies out of 14,137 sentences 
			in the Penn treebank tree-tagged corpus, the Wall Street Journal.
			The average frequency of each rule is 48 times in the corpus. 
		
			Of these rules, 
			we remove rules which occurs fewer times than the average frequency in the corpus,
			and then only 192 rules are left.

			These removed rules are almost for peculiar sentences and the left rules 
			are very general rules.
			We can show that our robust parser can compensate for lack of rules 
			using only 192 rules with the recovery mechanism and heuristics.

	\item	{\it Test set}\\
			First,  1,000 sentences are selected randomly from the WSJ corpus, which we have
			referred to in proposing the robust parser.
			Of these sentences, 410 are failed in normal parsing,
			and are processed again by the robust parser.
			To show the validity of these heuristics, 
			we compare the result of the robust parser using  heuristics  with one not using
			heuristics.
			Second, to show the adaptability of our robust parser,
	
			same experiments are carried out on
			1,000 sentences from the ATIS corpus in Penn treebank, which we haven't referred to 
			when we propose the robust parser.
			Among 1,000 sentences from the ATIS, 465 sentences are processed by the robust parser
			after the failure of the normal parsing.

	\item	{\it Parameter adjustment}\\
			We chose the best parameters of heuristics by executing several experiments. 
			
			\begin{tabbing}
			aa\=aaaaaaaaaa\=aa\=aaaaa\=aaaa\=aaaaaaaaaa\=aa\=aaaaa\=\kill
			\> $\alpha_{insertion}$ \> : \>  10.2 \>  \> $\beta_{insertion}$ \> : \> 15.0 \\
			\> $\alpha_{deletion}$ \> : \> 10.4   \>  \> $\beta_{deletion}$  \> : \> 20.0 \\
			\> $\alpha_{mutation}$ \> : \> 10.8\\
			\> $\delta_{1}$ \> : \> 0.01 \> \> $\delta_{2}$ \> : \> 5.0 \\
			\> $\delta_{3}$ \> : \> 1.0
			\end{tabbing}
\end{itemize}

\par
Accuracy is measured as the percentage of constituents in the test sentences which do not
cross any Penn treebank constituents \cite{ibm}.
Table~\ref{tab1} shows the results of the robust parser on WSJ.
In table~\ref{tab1}, {\it 5th, 6th} and  {\it 7th} raw mean that the percentage of sentences 
which have no crossing constituents, less than one crossing and less than two crossing respectively.
With heuristics, our robust parser can enhance the processing time and reduce the
number of edges. 
Also, the accuracy is improved from 72.8\% to 77.1\%  
even if the heuristics differentiate edges and prefer some edges.
It shows that the proposed heuristics is valid in parsing the real sentences.
The experiment says that
our robust parser with heuristics can recover perfectly
about 23 sentences out of 100 sentences which are just failed in normal parsing,
as the percentage of no-crossing sentences is about 23.28\%.

Table ~\ref{tab2} is the results of the robust parser on ATIS 
which we did not refer to before.
The accuracy of the result on ATIS is lower than WSJ because 
the parameters of the heuristics are  adjusted not by ATIS itself but  by WSJ.
However, the percentage of sentences with constituents crossing less than 2 is higher than the WSJ,
as sentences of ATIS are more or less simple.

        \begin{table*}
        \begin{center}
        \caption{The results of the robust parser on ATIS}
        \label{tab2}
        \renewcommand{\arraystretch}{1.1}
        \begin{tabular}{|l||l|l|}\hline
        \multicolumn{3}{|c|}{Experiment 2 : ATIS  465 sentences} \\\hline
                            &   with Heuristics             & without Heuristics\\\hline\hline
        Average sentence length    &   10.55 words (2-25 words)    & 10.55 words (2-25 words)    \\
        Average processing time &   8.68 sec                  & 71.98 sec                  \\
        Average number of edges &   12974.2                   & 25652.5                    \\
        Accuracy (\%)           &   68.5                    & 59.4                    \\
        no-crossing  sentences  &   26.02\%            & 13.28\%          \\
        \% of $\leq$ 1-crossing sentences&  47.10\%    & 36.06\%         \\
        \% of $\leq$ 2-crossing sentences&  66.24\% & 52.46\%        \\  \hline
        \end{tabular}
        \end{center}
        \end{table*}

The experimental results of our robust parser show high accuracy in recovery
even though 96\% of total rules are removed.
It is impossible to construct complete grammar rules in the real parsing system to
succeed in analyzing every real sentence.
So, parsing systems are likely to have extragrammatical sentences which cannot be analyzed
by the systems.
Our robust parser can recover these extragrammatical sentences
with 68 $\sim$ 77\% accuracy.

It is very interesting that parameters of heuristics reflect the characteristics of the test corpus.
For example, if people tend to  write sentences with inserted phrases, 
then the parameter $\beta_{insertion}$  must increase.
Therefore we can get better results if the parameter are fitted 
to the  characteristics of the corpus.

\section{Conclusion}

In this paper, we have presented the robust parser with the extended least-errors
recognition algorithm as the recovery mechanism.
This robust parser can easily be 
scaled up and applied to various domains
because this parser depends only on syntactic factors.
To enhance the performance of the robust parser for extragrammatical sentences, 
we proposed several heuristics.
The heuristics assign the error values to each error-hypothesis edge,
and  edges which has less error  values are processed first.
So, not all the generated edges are processed by the robust parser, 
but
the most plausible parse trees can be generated first.
The accuracy of the recovery in our robust parser is about 68\% $\sim$ 77\%.
Hence, this parser is suitable for systems in real application areas.

\par
Our short term goal is to propose an automatic method that can learn
parameter values of heuristics by analyzing the corpus.
We expect that automatically learned values of parameters can upgrade the performance of the parser.

\section*{Acknowledgement}

This work was supported(in part) by Korea Science and Engineering Foundation(KOSEF)
through Center for Artificial Intelligence Research(CAIR), the Engineering Research Center(ERC)
of Excellence Program.

\end{document}